%% file: main.tex
\documentclass[10pt,twocolumn,letterpaper]{article}

\usepackage[pagenumbers]{cvpr} 
\usepackage{amsmath}
\usepackage{multicol}
\usepackage{multirow}
\usepackage{fancyhdr}
\usepackage{caption}
\usepackage{subcaption}
\input{preamble}

\definecolor{cvprblue}{rgb}{0.21,0.49,0.74}
\NewDocumentCommand{\yizhu}
{ mO{} }{\textcolor{red}{\textsuperscript{\textit{Yizhu}}\textsf{\textbf{\small[#1]}}}}

\usepackage[pagebackref,breaklinks,colorlinks,citecolor=cvprblue]{hyperref}
\usepackage{array}
\usepackage{tabularx}
\def\paperID{10849} 
\def\confName{CVPR}
\def\confYear{2024}

\title{DiffPoint: Single and Multi-view Point Cloud Reconstruction with ViT Based Diffusion Model}

\author{Yu Feng\\
Shanghai Key Laboratory of Data Science, \\School of Computer Science, \\Fudan University, \\China\\
{\tt\small yufeng9819@gamil.com}
\and
Xing Shi\\
Alibaba Group\\
China\\
{\tt\small shubao.sx@alibaba-inc.com}
\and
Mengli Cheng\\
Alibaba Group\\
China\\
{\tt\small mengli.cml@alibaba-inc.com}
\and
Yun Xiong\\
Shanghai Key Laboratory of Data Science, \\School of Computer Science, \\Fudan University, \\China\\
{\tt\small yunx@fudan.edu.cn}
}

\begin{document}
\maketitle
\input{sec/0_abstract}    
\input{sec/1_introduction}
\input{sec/2_related_work}
\input{sec/3_method}
\input{sec/4_experiments}
\input{sec/5_conclusion}
{
    \small
    \bibliographystyle{ieeenat_fullname}
    \bibliography{main}
}

\input{sec/X_suppl}

\end{document}

%% file: preamble.tex
\usepackage[dvipsnames]{xcolor}


%% file: sec/0_abstract.tex
\begin{abstract}
    As the task of 2D-to-3D reconstruction has gained significant attention in various real-world scenarios, it becomes crucial to be able to generate high-quality point clouds. Despite the recent success of deep learning models in generating point clouds, there are still challenges in producing high-fidelity results due to the disparities between images and point clouds. While vision transformers (ViT) and diffusion models have shown promise in various vision tasks, their benefits for reconstructing point clouds from images have not been demonstrated yet. In this paper, we first propose a neat and powerful architecture called DiffPoint that combines ViT and diffusion models for the task of point cloud reconstruction. At each diffusion step, we divide the noisy point clouds into irregular patches. Then, using a standard ViT backbone that treats all inputs as tokens (including time information, image embeddings, and noisy patches), we train our model to predict target points based on input images. We evaluate DiffPoint on both single-view and multi-view reconstruction tasks and achieve state-of-the-art results. Additionally, we introduce a unified and flexible feature fusion module for aggregating image features from single or multiple input images. Furthermore, our work demonstrates the feasibility of applying unified architectures across languages and images to improve 3D reconstruction tasks.
\begin{figure*}[!htbp]
\centering
\includegraphics[width=1\linewidth]{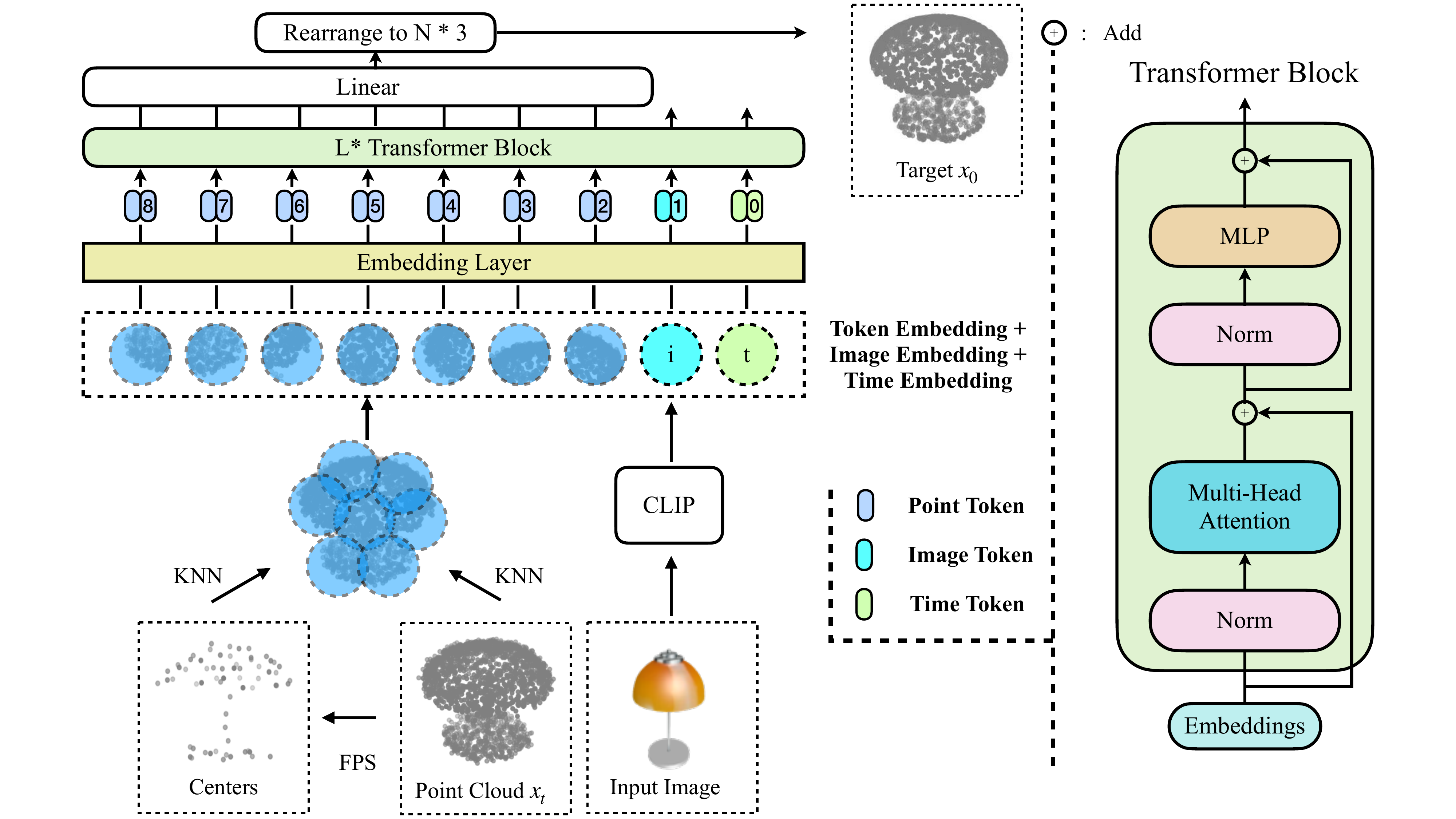}
\caption{\textbf{Overall architecture of our DiffPoint.} DiffPoint is built on a standard ViT backbone, treating all inputs including the time image embedding and noisy point cloud patches as tokens and targeting ground truth data $x_0$ for prediction.}
\label{fig:method}
\end{figure*}

\end{abstract}

%% file: sec/1_introduction.tex
\section{Introduction}
\label{sec:intro}

In recent years, the field of 3D computer vision has seen a surge of interest in converting 2D images into 3D models, a process known as 2D-to-3D reconstruction. This technique aims to generate a three-dimensional representation of an object from one or more two-dimensional images. Its applications are widespread, including robotics \cite{Maboudi2022ARO}, autonomous driving \cite{Huang2022MultimodalSF,Kumar2022SurroundviewFC}, virtual reality \cite{Lunghi2017AnRB}. However, the main challenge of this task lies in establishing accurate feature matching between the 2D images and the 3D output, primarily due to their inherent differences. To address this challenge, deep learning models for 3D reconstruction have made significant advancements in terms of quality and flexibility. Existing models can be broadly categorized into three types based on the explicit representation of 3D data: voxel-based models \cite{Choy20163DR2N2AU,Xie2019Pix2VoxC3,Xie2020Pix2VoxMC, Yagubbayli2021LegoFormerTF}, mesh-based models \cite{Wen2022Pixel2Mesh3M,Wang2018Pixel2MeshG3,Pan2019DeepMR}, and point-based models \cite{Wen20223DSR,Fan2017APS}. While voxel-based representations suffer from high memory costs at high resolutions and meshes struggle to model inner or irregular structures, point clouds offer lightweight storage consumption and are capable of representing various complex shapes. In this paper, we focus on point-based reconstruction.
\par

Though these methods achieve some success to some extent, the 2D-to-3D reconstruction task is still suffering from the following problem: how to find an effective way to reduce the disparities between images and 3D shapes for reconstructing accurate results. Most previous methods \cite{Choy20163DR2N2AU,Xie2019Pix2VoxC3,Xie2020Pix2VoxMC,Yagubbayli2021LegoFormerTF,Wen2022Pixel2Mesh3M,Wang2018Pixel2MeshG3,Pan2019DeepMR} follow the standard encoder-decoder structure, in which image features are treated as additional information and limited semantic information can be conveyed to the 3D decoder. As a result, these methods fail to reconstruct satisfactory results. Driven by scaling models on large datasets, diffusion models \cite{SohlDickstein2015DeepUL,Ho2020DenoisingDP,Song2020ScoreBasedGM} have shown great success in various tasks \cite{Nichol2022GLIDETP,Dhariwal2021DiffusionMB,Tashiro2021CSDICS,Kong2021DiffWaveAV,Kim2022GuidedTTSAD, Radford2021LearningTV}. Diffusion models have emerged as highly effective generative models, showcasing exceptional performance in not only image synthesis \cite{Ramesh2022HierarchicalTI, Chang2023MuseTG, Dhariwal2021DiffusionMB, Preechakul2021DiffusionAT,Rombach2021HighResolutionIS} but also the generation of 3D shapes based on point clouds \cite{Luo2021DiffusionPM,Zeng2022LIONLP,Zhou20213DSG,Mo2023DiT3DEP}. Diffusion models show potential to be adopted for 3D reconstruction. However, existing point-based diffusion models cannot fully solve the problem above. These methods, implemented with MLP \cite{Luo2021DiffusionPM} and PVCNN \cite{Zeng2022LIONLP,Zhou20213DSG}, also treat image embeddings as additional information and fail to fully explore visual information. Moreover, DPM based on MLP struggles to reconstruct complex structure shapes. For Point-Voxel CNN (PVCNN) based diffusion models \cite{Zeng2022LIONLP,Zhou20213DSG}, they require huge memory and computation costs to implement voxel convolution operation. On the other hand, vision transformers (ViT) \cite{Dosovitskiy2020AnII} have shown promise in various vision tasks due to their flexible backbone. However, directly applying ViT to point cloud reconstruction may also suffer from the following problem: while ViT is excellent at capturing global context, it may struggle with capturing fine-grained local details in images. Point cloud data demands a granular understanding of individual point positions and features, which the traditional attention mechanisms in ViT, optimized for grid-like image structures, might not adequately address. In a word, existing point-based diffusion models and ViT cannot be directly solve the problem above.
\par

In this paper, we combine diffusion model and ViT to propose DiffPoint, a neat and flexible ViT-based diffusion architecture for point cloud reconstruction. The diffusion model excels in capturing local intricacies and dependencies within data, providing a complementary capability to ViT's global context awareness. Directly extending ViT and diffusion models to point cloud is, however, technically highly nontrivial: unlike images, point clouds consist of discrete points in 3D space, making it impossible to divide them into patches directly. To address this issue, we follow the design methodology of PointMAE \cite{Pang2022MaskedAF} and Point-BERT \cite{Yu2021PointBERTP3}, where point cloud are divided into irregular patches through Farthest Point Sampling (FPS) and K-Nearest Neighborhood (KNN) algorithm. These divided patches are then encoded into token embeddings using a lightweight encoder called PointNet \cite{Qi2016PointNetDL}. Inspired by CLIP's ability to learn robust representations capturing both semantics and style in images, we leverage these representations for point cloud reconstruction. The input images are encoded with CLIP into image embeddings. Similar to U-ViT \cite{Bao2022AllAW}, we treat all inputs including the time, image embedding, noisy points patches as tokens that interact within the same feature space. To sum up, combining diffusion models with ViT may address the problem above: \textbf{a)} The ViT backbone in diffusion models helps in creating a more expressive and context-aware representation of features in which hierarchical attention mechanisms of ViT capture both global and local information effectively; \textbf{b)} ViT's token-based approach allows for the assimilation of diverse information, including time, image embeddings, and patches, effectively bridging the gap between different modalities. Besides, we introduce a unified multi feature aggregation module for single and multi-view tasks ensuring consistency in the model architecture. 
\par

We evaluate our DiffPoint in both single and multi-view reconstruction tasks. Our results on popular benchmark ShapeNet \cite{Chang2015ShapeNetAI} suggests that DiffPoint is capable of reconstruct high-fidelity point clouds, outperforming multiple state-of-the-art methods. Furthermore, we extend DiffPoint to a more complicated and diverse dataset OBJAVERSE-LVIS \cite{Deitke2022ObjaverseAU}, demonstrating the powerful modeling capabilities of DiffPoint and the potential to migrate to large-scale 3D datasets.
\par

The main contributions of this paper are summarized as follows:
\begin{itemize}
    \item We explore 3D point cloud reconstruction and introduce a novel architecture, DiffPoint, which combines diffusion models with ViT. DiffPoint enhances image feature representation and connects the 2D and 3D domains. Our work marks the first use of ViT-based diffusion models for 3D point cloud reconstruction. 
    \item We propose a unified module for aggregating multiple features, which can be used for single and multi-view reconstruction tasks ensuring consistency in the model architecture for both tasks.
    \item Our model demonstrates state-of-the-art performance in reconstructing 3D shapes from both single-view and multi-view perspectives, with exceptional modeling capabilities confirmed through additional experiments on OBJAVERSE-LVIS. 
\end{itemize}

%% file: sec/2_related_work.tex
\section{Related Work}
\label{sec:rel}
\noindent \textbf{2D-to-3D Reconstruction.} 
 Early 2D-to-3D reconstruction tasks \cite{Fan2017APS, Groueix2018AtlasNetAP, Mescheder2019OccupancyNL, Wang2018Pixel2MeshG3, Wen20223DSR} have gained attention in the 3D computer vision community. Reconstructing 3D shapes from 2D images involves establishing a connection between the shape of 3D data and 2D images. It can be seen as a multi-modal learning problem that requires a large amount of supervised information in the form of 3D data. There are two ways to categorize 2D-to-3D reconstruction tasks based on data representation and the number of input images. There are two common categorizations that distinguish based on the number of input images: single-view 3D shape reconstruction \cite{Groueix2018AtlasNetAP,Mescheder2019OccupancyNL,Wang2018Pixel2MeshG3,Wang2018AdaptiveO} and multi-view 3D shape reconstruction \cite{Wen2022Pixel2Mesh3M,Xie2020Pix2VoxMC,Xie2019Pix2VoxC3,Yagubbayli2021LegoFormerTF}.
 \par
 
Categorized by output representation, early works based on explicit representation can be divided into voxel-based, mesh-based, or point-based methods. Voxel-based methods typically use 3D convolutional neural networks \cite{Wu20153DSA} and leverage the CNN structure in both 2D and 3D domains. However, voxel-based methods suffer from the cubic growth of voxel data and are limited to low resolution results \cite{Choy20163DR2N2AU,Xie2019Pix2VoxC3,Xie2020Pix2VoxMC,Yagubbayli2021LegoFormerTF}. Mesh-based methods usually capture semantic information in 2D images using a mesh deformation network \cite{Wen2019Pixel2MeshM3,Wang2018Pixel2MeshG3,Pan2019DeepMR}, but they lack the ability to represent inner or irregular structures. In contrast, point-based methods such as DiffPoint reconstruct point cloud data from 2D images with lower memory usage and more detailed inner structure information \cite{Wen20223DSR,Fan2017APS}. On the other hand, implicit representation based methods like Neural radiance fields (NeRF) optimize 5D neural radiance fields through differentiable rendering using image pixel colors \cite{Mildenhall2020NeRF,Liu2020NeuralSV,Barron2021MipNeRFAM,Yu2020pixelNeRFNR,Bian2022NoPeNeRFON,Wynn2023DiffusioNeRFRN,Muller2022DiffRFR3}. These approaches differ from our model in terms of training data and optimizer target. Additionally, these approaches require dense input images for compelling results and their performance degrades when only a few views are available.
\newline
 
\noindent \textbf{Point Based Diffusion Model.}
The Denoising Diffusion Probabilistic Model (DDPM) \cite{Ho2020DenoisingDP, SohlDickstein2015DeepUL} has gained significant momentum in recent years. As a new type of generative model, it has demonstrated superior performance in various generation tasks, including image generation \cite{Ho2020DenoisingDP, Dhariwal2021DiffusionMB}, text-to-image \cite{Ramesh2022HierarchicalTI, Chang2023MuseTG, Dhariwal2021DiffusionMB, Preechakul2021DiffusionAT,Rombach2021HighResolutionIS,Bao2022AllAW}, time series imputation \cite{Tashiro2021CSDICS}, and text-to-speech synthesis \cite{Popov2021GradTTSAD}. In the field of image generation, U-Net based diffusion models continue to dominate while U-ViT \cite{Bao2022AllAW} have been proposed to explore the use of ViT for image generation with diffusion models. Regarding the point generation area, point-based diffusion models \cite{Luo2021DiffusionPM,Zeng2022LIONLP,Zhou20213DSG,Mo2023DiT3DEP} are introduced to address unconditional or class-conditional point generation problems. However, due to limitations in the backbone of diffusion models, DPM \cite{Luo2021DiffusionPM} based on MLP struggles with modeling large-scale shapes. Similarly, LION \cite{Zeng2022LIONLP} and PVD \cite{Zhou20213DSG}, which are based on PVCNN, require significant memory and computation costs for generating high-resolution shapes. A related work called DiT-3D \cite{Mo2023DiT3DEP} utilizes ViT for point diffusion models but does not explore the 3D reconstruction problem like our DiffPoint does. Additionally, we differ in how to handle input noisy point clouds as it would incur additional computation costs to convert them into voxels and may limit shape resolution in real-world application scenarios.
\newline
 
\noindent \textbf{Transformers.}
Transformers \cite{Vaswani2017AttentionIA} have demonstrated exceptional performance in various domains, including NLP \cite{Radford2018ImprovingLU,Brown2020LanguageMA,Raffel2019ExploringTL,Devlin2019BERTPO} and images \cite{Dosovitskiy2020AnII,He2021MaskedAA,Liu2021SwinTH,Yuan2021TokenstoTokenVT}. ViT \cite{Dosovitskiy2020AnII} is one of the earliest methods to apply transformers to image areas by dividing images into patches. Recently, PointTransformer \cite{Engel2020PointT}, MaskPoint \cite{Liu2022MaskedDF}, and PointMAE \cite{Pang2022MaskedAF} have attempted to incorporate transformers into 3D point cloud analysis. However, there has been limited exploration of combining ViT and diffusion models for point cloud data. To the best of our knowledge, along with \cite{Mo2023DiT3DEP}, we are the first to investigate the use of ViT-based diffusion models for 3D point cloud analysis.
\newline

%% file: sec/3_method.tex
\section{Method}
\label{sec:method}
\subsection{Background}
\noindent \textbf{Diffusion Models.} Diffusion models \cite{Ho2020DenoisingDP, SohlDickstein2015DeepUL, Song2020ScoreBasedGM} are a class of promising generative models that introduce noise to data and then reverse this process to generate data from the noise. This procedure, known as the forward process or noise-infection procedure, is formalized as a Markov chain: $q(X^{1},...,X^{T}|X^{0})=\prod_{t=1}^{T}{q(X^{t}|X^{t-1})}$.
It adds small Gaussian noise to points at the previous time step as follows: $q(X^{t}|X^{t-1})=\mathcal{N}(X^{t};{\sqrt{1-\beta_{t}}}{X^{t-1}}, \beta_{t}\boldsymbol{I}),$ where $\beta_{t}$ is a small positive constant and $\boldsymbol{I}$ is the identity matrix with same dimension as $X^{t-1}$.
\par

The reverse process is defined as the conditional distribution $p_\theta(X^{(0:T-1)}|X^{T}, c)$, where $c$ is input images features encoded by the image encoder. The whole reverse process can be factorized into multiple transitions based on the Markov chain:
\begin{equation}
  p_\theta{(X^{0},...,X^{T-1}|X^{T},c)}=\prod^{T}_{t=1}{p_\theta(X^{t-1}|X^{t},c)}.
  \label{eq:reverse_trans}
\end{equation}\noindent
In the reverse process, the Markov kernel is parameterized as: $p_\theta(X^{t-1}|X^{t},c)=\mathcal{N}({X^{t-1};\mu_\theta{(X^t,t),\sigma_{\theta}^2{(X^t,t)}\boldsymbol{I}})}$,
where $\mu_\theta{(X^t,t)}$ and $\sigma_{\theta}^2{(X^t,t)}$ are mean and variance of the reverse process respectively. By utilizing the reverse transitions $p_\theta{(X^{t-1} | X^{t},c)}$, the latent variables are gradually returned to the point cloud that aligns with the diffusion time-step and image condition.
\par

For training objective, we optimize the simple objective proposed by DDPM \cite{Ho2020DenoisingDP}:
\begin{equation}
\begin{aligned}
    \underset{\theta}{\text{min}}L(\theta) := \underset{\theta}{\min}\mathbb{E}||X^0-\epsilon_{\theta}(\sqrt{\Bar{\alpha}_t}X^0+\sqrt{1-\Bar{\alpha}_t}\epsilon,t,c)||^{2}.
  \label{eq:trian_loss}
\end{aligned}
\end{equation}
In DiffPoint, unlike DDPM which uses $\epsilon$-prediction and mean squared error as the loss function to predict noise, we predict the target points $X^0$, which can be interchangeably used for predicting noise \cite{Ramesh2022HierarchicalTI}. We utilize the Chamfer distance (CD) as our loss function. 
\subsection{Overview}
The proposed DiffPoint aims to reconstruct the point cloud of an object from single or multiple images. This paper presents a neat and efficient scheme for point cloud reconstruction using a ViT-based diffusion model. Figure \ref{fig:method} depicts the overall approach of our method, DiffPoint. In each diffusion step, noisy points undergo initial processing through a splitting and embedding module. Subsequently, input images (single or multi-view) are processed using pretrained CLIP and feature aggregation modules. Finally, a standard ViT-based backbone is employed to predict ground truth data.

\subsection{Point Cloud Splitting and Embedding}
Generally, ViT processes an image as a sequence of flattened patches. However, consisting of discrete points, point clouds cannot be directly divided into regular patches like images. To align with the standard ViT backbone, we adopt the approach used in Point-BERT \cite{Yu2021PointBERTP3} and PointMAE \cite{Pang2022MaskedAF} , which involves dividing the point cloud into irregular point patches using Farthest Point Sampling (FPS) \cite{Qi2017PointNetDH} and the K-Nearest Neighborhood (KNN) algorithm. At each diffusion step, $s$ centers are sampled for point patches through FPS. Here, $s$ represents the number of point patches. Once center points are determined, patches are generated by gathering $k$ nearest neighbors from noisy points $X^{t}$. Ultimately, noisy points $X^{t}$ are divided into $s$ patches with a patch size of $k$. As argued in \cite{Pang2022MaskedAF}: simply linear embedding against the principle of permutation invariance \cite{Qi2016PointNetDL}. Therefore, we employ PointNet \cite{Qi2016PointNetDL} to encode each point patch into feature embeddings that can be processed by the ViT backbone.

\begin{figure}[!htbp]
\centering
\includegraphics[width=1\linewidth]{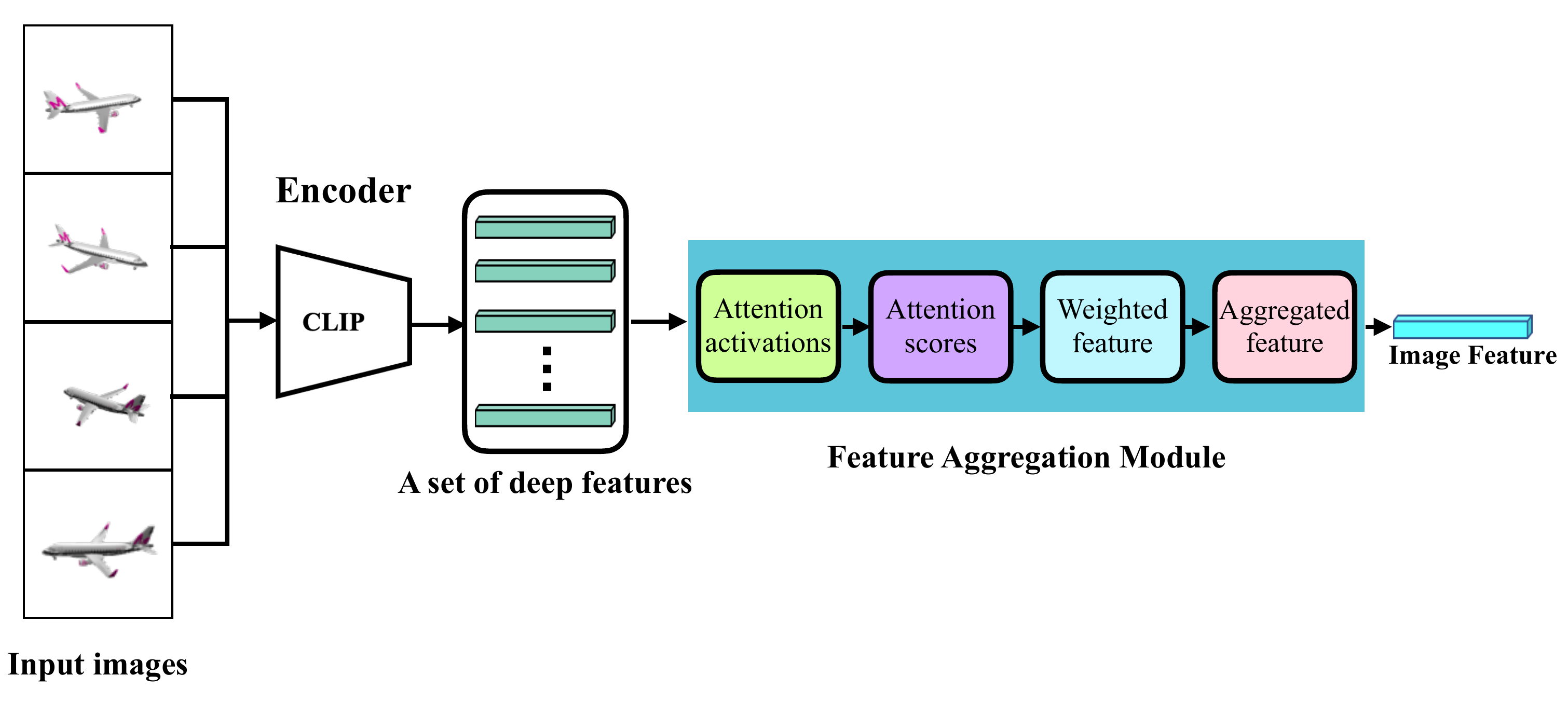}
\caption{Illustration of feature aggregation module. Multi image features encoded by CLIP are aggregated through attention mechanism.}
\label{fig:feature_aggregation_module}
\end{figure}

\subsection{Image Embedding and Feature Aggregation}
The remarkable success of CLIP \cite{Radford2021LearningTV} has demonstrated its ability to learn image representations. Naturally, we adopt the pretrained CLIP with a ViT-based visual backbone to encode the input image. Moreover, for multi-view reconstruction, models need to aggregate features from different views of an object. Taking inspiration from the flexible approach \cite{Yang2018RobustAA} of aggregating multiple image features into one embedding, we propose a feature aggregation module that utilizes self-attention mechanism to aggregate multi-image features. This module can receive features not only from multi-view inputs but also from single images, resulting in a neat and consistent model framework.

\begin{table*}[htbp!]
\centering
\caption{Single-view reconstruction on ShapeNet dataset in terms of CD $\times10^2$ (lower is better) and F-score($\tau$) $\times10^2$ (larger is better and $\tau=10^{-3}$).}
\label{tab:exp_single}
\scalebox{0.65}{
 \begin{tabular}{c|ccc ccc c |ccc ccc} 
\toprule[2pt]
    \multirow{2}{*}{\textbf{Methods}} & \multicolumn{7}{c|}{CD$\downarrow$}& \multicolumn{6}{c}{F-Score($\tau$)$\uparrow$} \cr
    & 3DR2N2 & PSGN & Pixel2Mesh & AtlasNet & OccNet &3DAttriFlow&\textbf{DiffPoint-S} & 3DR2N2 & PSGN & Pixel2Mesh  & OccNet &3DAttriFlow&\textbf{DiffPoint-S}\\
\midrule[2pt]
\midrule[2pt]
      Plane &4.94 &2.78 &5.36 &2.60 &3.19 &2.11 &\textbf{1.69} &85.05 &\textbf{88.24} &77.93 &79.30 &81.55 &87.95 \\
      Bench &4.80 &3.73 &5.14 &3.20 &3.31 &2.71&\textbf{1.72} &81.12 &80.34 &75.80 &76.60 &72.11 &\textbf{88.00}\\
      Cabinet &4.25 &4.12 &4.85 &3.66 &3.54 &2.66 &\textbf{2.30} &68.08 &68.94 &70.07 &74.78 &71.49 &\textbf{78.26}  \\
      Car &4.73 &3.27 &4.69 &3.07 &3.69 &2.50 &\textbf{2.39} &72.40 &\textbf{88.01} &76.76 &77.55 &74.30 &75.83  \\
      Chair &5.75 &3.27 &5.77 &4.09 &4.08 &3.33&\textbf{2.12} &68.38 &63.65 &61.92 &66.67 &59.40 &\textbf{79.52}  \\
      Display &5.85 &4.68 &5.28 &4.16 &4.84 &3.60 &\textbf{2.43} &62.82 &61.98 &65.78 &62.76 &58.56 &\textbf{74.38}  \\
      Lamp &10.64 &4.74 &6.87 &4.98 &7.55 &4.55 &\textbf{1.99} &62.25 &58.78 &62.47 &56.91 &51.43 &\textbf{80.95} \\
      Speaker &5.96 &5.60 &6.17 &4.91 &5.47 &4.16 &\textbf{2.86} &57.50 &53.74 &56.12 &58.54 &52.10&\textbf{73.71}\\
      Rifle &4.02 &5.62 &4.21 &2.20 &2.97 &1.94 &\textbf{1.82} &86.90 &\textbf{91.12} &83.38 &84.53 &84.84 &87.08 \\
      Sofa &4.72 &2.53 &5.34 &3.80 &3.97 &3.24 &\textbf{2.53} &71.97 &71.11 &70.28 &68.10 &61.32 &\textbf{72.62}  \\
      Table &5.29 &4.44 &5.13 &3.36 &3.74 &2.85 &\textbf{2.31} &72.47 &74.75 &70.43 &74.62 &70.32 &\textbf{77.18} \\
      Phone &4.37 &3.81 &4.22 &3.20 &3.16 &2.66 &\textbf{2.37} &83.27 &82.72 &82.56 &\textbf{86.79} &74.71 &77.70 \\
      Vessel &5.07 &3.84 &5.48 &3.40 &4.43 &2.96 &\textbf{1.98} &75.46 &82.41 &73.39 &68.96 &67.16&\textbf{82.62} \\
\midrule[2pt]
      Mean &5.41 &4.07 &5.27 &3.59 &4.15 &3.02 & \textbf{2.19} &72.90 &74.29 &71.30 &72.00 &67.64&\textbf{79.67}\\
\bottomrule[2pt]
\end{tabular}}
\end{table*}

\subsection{Diffusion Backbone}
As shown in Figure \ref{fig:method}, our DiffPoint's backbone is built entirely on the standard ViT. It treats all inputs, including time, image, and noisy point patches, as tokens. The prediction network $\epsilon_{\theta}(X^t,t,c)$ consists of standard Transformer blocks. Following \cite{Yu2021PointBERTP3}, we apply MLP positional embedding to each token before adding them to the Transformer blocks. In the last layer of the backbone, the output aims to predict the target point cloud $X^0$ in coordinate space. We simply use a fully connected layer as a project embedding from the last transformer block to the output with dimensions matching those of $X^0$. 

%% file: sec/4_experiments.tex
\section{Experiments}
\label{sec:exp}

In this section, we assess the effectiveness of DiffPoint by evaluating single and multi-view 3D reconstruction tasks. We implement two versions of our proposed framework for both tasks: \textbf{DiffPoint-S} for single-view and \textbf{DiffPoint-M} for multi-view. Specifically, we compare our method with previous state-of-the-art models through quantitative and qualitative comparisons. To further analyze its modeling capability, we extend DiffPoint to train on the OBJAVERSE-LVIS dataset \cite{Deitke2022ObjaverseAU}. The ablation studies focus on assessing the effectiveness of DiffPoint's architecture in the multi-view 3D reconstruction task
\begin{figure*}[!htbp]
\centering
\includegraphics[width=0.8\linewidth]{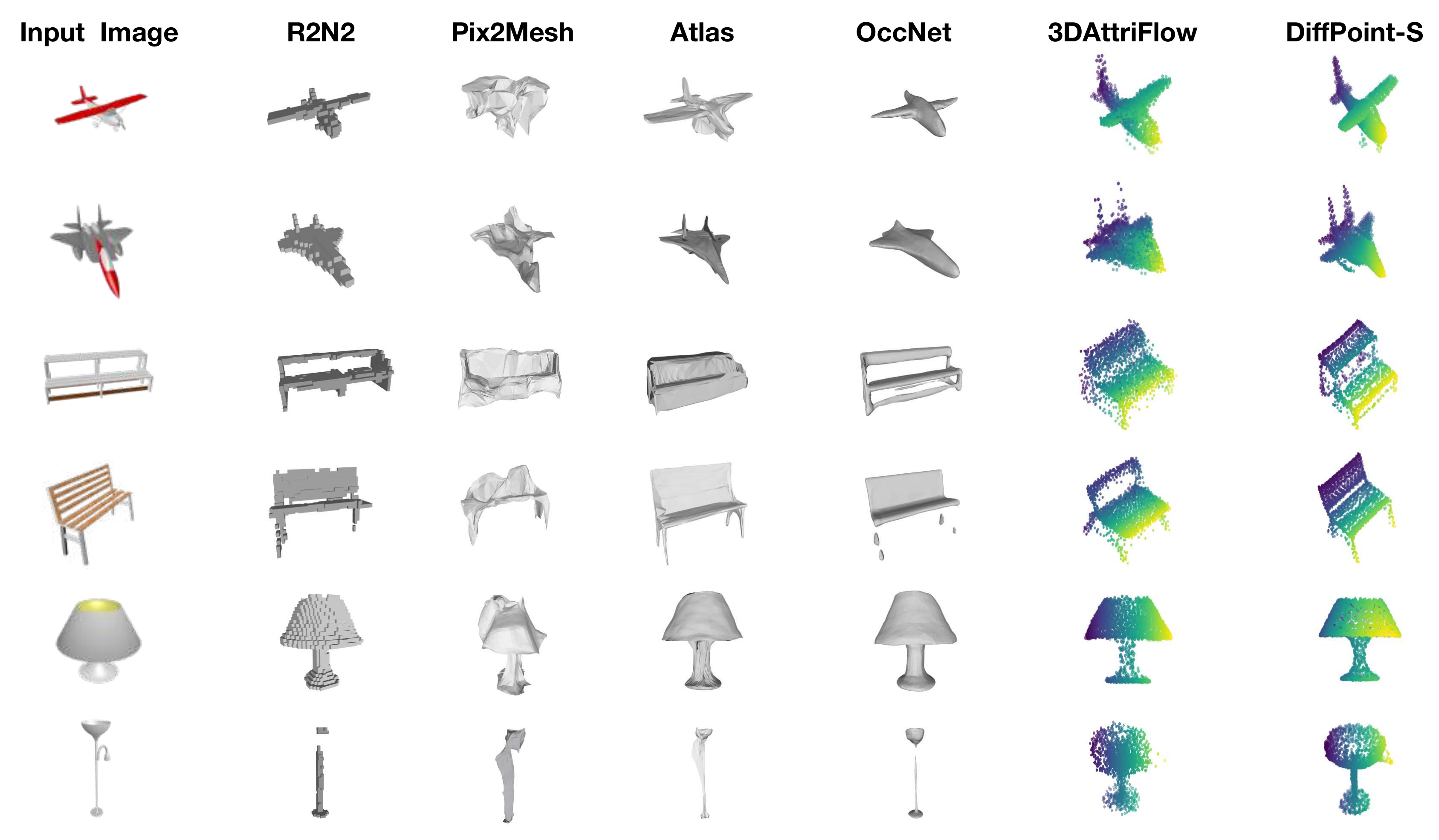}
\caption{\textbf{Single View 3D Reconstruction.}The input image is shown in the first column, the other columns show the results for our method compared to various baselines.}
\label{fig:single_vis}
\end{figure*}

\subsection{Experimental setup}
\noindent \textbf{Baselines.} For single-image 3D reconstruction, we compare our method with state-of-the-art baselines, including voxel-based (3D-R2N2 \cite{Choy20163DR2N2AU}), mesh-based (Pixel2Mesh \cite{Wang2018Pixel2MeshG3} and AtlasNet \cite{Groueix2018AtlasNetAP}), and point-based (PSGN \cite{Fan2017APS} and 3DAttriFlow \cite{Wen20223DSR}). In the realm of multi-image 3D reconstruction, we assess against voxel-based methods (3D-R2N2, Pix2Vox++ \cite{Xie2020Pix2VoxMC}, LegoFormer \cite{Yagubbayli2021LegoFormerTF}) and the mesh-based approach Pixel2Mesh++. Additionally, we extend 3DAttriFlow for multi-view reconstruction, denoted as 3DAttriFlow-M, by averaging features from multiple views, as multi-view point cloud reconstruction is a novel exploration. All baseline methods utilize their respective publicly available code repositories.
\par

\noindent \textbf{Datasets and evaluation metric.}
For the single-view reconstruction task, we adhere to the settings used in previous work \cite{Wen20223DSR} and only evaluate our DiffPoint on ShapeNet dataset \cite{Chang2015ShapeNetAI}. It has 43,783 mesh objects covering 13 categories. Following OccNet \cite{Mescheder2019OccupancyNL}, we split the dataset into training, validation, and test sets with the ratio of 70\%, 10\%, and 20\%, respectively. When calculating metrics, we follow AtlasNet \cite{Groueix2018AtlasNetAP} to uniformly sample 30k points on the mesh surface of the 3D object as ground truth for testing. Regarding metrics, we evaluate our reconstructed points by comparing them to ground truth using L1 Chamfer distance (CD), which is defined as:
\begin{equation}
\begin{aligned}
    L_{CD}(P^r,P^g)=&\frac{1}{2N}\sum_{p^r\in{P^r}}{\underset{p^g\in{P^g}}{\min}||p^r-p^g||_2} 
    \\& + \frac{1}{2N}\sum_{p^g\in{P^g}}{\underset{p^r\in{P^r}}{\min}||p^t-p^r||_2},
  \label{eq:l1_loss}
\end{aligned}
\end{equation}
In addition, we use the F-score following \cite{Wang2018Pixel2MeshG3} to evaluate the distance between the object surface and it is defined as the harmonic mean between precision and recall. We set $\tau = 10^{-3}$ as suggested by \cite{Tatarchenko2019WhatDS}.
\par

For the multi-view reconstruction task, we maintain consistent experimental setup as the single-view reconstruction experiments. We evaluate on ShapeNet and use CD and F-score as the metric. In the 3D reconstruction task, one of the main concerns is designing a model with high generalization capability. However, ShapeNet, which is commonly used for training, only contains a limited number of objects with simple structures. Motivated by this limitation, we evaluate our DiffPoint on a challenging real-world dataset called OBJAVERSE-LVIS. This dataset consists of approximately 44,834 objects from over 1000 categories that have complex structures and are obtained from various real-world sources. Since OBJAVERSE-LVIS dataset publicly provides 3D data in GLB format only, it cannot be directly applied to point-based 3D reconstruction. Therefore, we preprocess the dataset to obtain point cloud data and generate 24 rendered images for each object similar to ShapeNet's format.
\par

\noindent \textbf{Implementation Details.}
Our models use PyTorch \cite{Paszke2017AutomaticDI} and are trained on Nvidia V100 using the Adamw \cite{Loshchilov2017DecoupledWD} optimizer with a batch size of 128. The learning rate is set to 2e-4. At each iteration, the input images are randomly sampled from 24 views. Both tasks differ in the number of sampled images. After being preprocessed by CLIP, the original $137 \times 137$ images are resized to $224 \times 224$. During our experiment, we found it difficult for DiffPoint-S to converge when trained on all categories. To make training easier, DiffPoint-S is trained and tested per category on ShapeNet dataset. For the backbone of DiffPoint-S, we set the number of ViT blocks to 17 and token embedding dimension to 384 for each category.  \par
Unlike DiffPoint-S, DiffPoint-M is trained on all categories of ShapeNet with more image views as input. The training process uses a fixed number of 5 input views. When evaluating the metrics for other baselines, we also maintain the same number of images as input. To handle larger-scale data, we set the depth of ViT to 19 and the hidden size to 512. It's important to note that both DiffPoint-S and DiffPoint-M divide noisy point clouds with 2048 points into 64 patches, each containing 32 points.

\begin{table*}[h]
\centering
\caption{Multi-view reconstruction on ShapeNet dataset in terms of CD $\times10^2$ (lower is better) and F-score($\tau$) $\times10^2$ (larger is better and $\tau=10^{-3}$).}
\label{tab:exp_multi}
\scalebox{0.6}{
 \begin{tabular}{c|ccc ccc |ccc ccc} 
\toprule[2pt]
    \multirow{2}{*}{\textbf{Methods}} & \multicolumn{6}{c|}{CD$\downarrow$}& \multicolumn{6}{c}{F-Score($\tau$)$\uparrow$} \cr
    & 3DR2N2 & Pix2Voxel & LegoFormer & Pixel2Mesh++ &3DAttriFlow-M&\textbf{DiffPoint-M} & 3DR2N2 & Pix2Voxel & LegoFormer & Pixel2Mesh++ &3DAttriFlow-M&\textbf{DiffPoint-M}\\
\midrule[2pt]
\midrule[2pt]
      Plane &4.17 &2.77 &2.95 &4.29  &1.85 &\textbf{1.69} &51.11 &69.89 &64.51 &52.18 &85.48 &\textbf{88.30} \\
      Bench &4.90 &3.33 &3.18 &5.11 &2.29 &\textbf{1.86} &42.88 &58.81&59.81 &45.37 &79.65 &\textbf{85.27}\\
      Cabinet &3.93 &3.36 &3.54 &5.65 &2.45 &\textbf{2.26} &49.52 &57.11&55.15 &33.99 &\textbf{79.52} &72.88  \\
      Car &3.29 &2.63 &2.82 &5.08 &2.30 &\textbf{2.26} &56.67 &67.42 &63.33 &35.74&\textbf{80.36} &77.19  \\
      Chair &4.87 &3.83 &4.15 &6.02 &2.78 &\textbf{2.09} &38.08 &48.62 &44.68 &31.39&68.98 &\textbf{79.70}  \\
      Display &5.14 &3.87 &3.84 &4.93 &2.87 &\textbf{2.22} &36.01 &49.59 &50.41 &44.26 &68.47 &\textbf{77.66}  \\
      Lamp &8.06 &6.80 &6.74 &8.43 &4.08 &\textbf{1.94} &33.22 &40.65 &39.01 &29.39 &57.35 &\textbf{82.79} \\
      Speaker &5.64 &4.87 &4.94 &7.03 &3.48 &\textbf{2.60} &36.13 &41.74 &41.20 &27.13 &60.84 &\textbf{68.07}\\
      Rifle &3.45 &2.58 &2.61 &3.12 &1.75 &\textbf{1.64} &61.46 &71.79 &70.63 &71.93 &87.84 &\textbf{88.99} \\
      Sofa &5.01 &3.68 &3.50 &5.36 &2.74 &\textbf{2.45} &39.59 &52.65 &53.93 &34.99 &71.00 &\textbf{71.38}  \\
      Table &4.68 &3.99 &4.04 &5.76 &2.55 &\textbf{2.02} &42.83 &50.34 &49.65 &37.16 &75.09 &\textbf{80.96} \\
      Phone &3.68 &2.71 &2.63 &4.05 &2.11 &\textbf{1.98} &60.85 &70.94 &71.81 &57.11 &83.02 &\textbf{83.40} \\
      Vessel &5.02 &3.84 &3.84 &4.61 &2.55 &\textbf{2.22} &44.22 &54.46 &53.44 &49.63 &73.47 &\textbf{78.90} \\
\midrule[2pt]
      Mean &4.82 &3.71 &3.75 &5.34 &2.58 &\textbf{2.10} &45.58 &56.46 &55.19 &42.32 &74.69 &\textbf{79.65}\\
\bottomrule[2pt]
\end{tabular}}
\end{table*}

\subsection{Single View Reconstruction}
\par

\noindent \textbf{Quantitative comparison.}
In Table \ref{tab:exp_single}, we present the performance of various single view baselines. Since we adopt the testing setting of 3DAttriFlow, we utilize the results from 3DAttriFlow for CD. Overall, our DiffPoint-S, which treats noisy point patches as input tokens, outperforms previous methods in terms of CD for each semantic category. We calculate F-score for other baselines (excluding 3DAttriFlow) following OccNet's implementation and DiffPoint-S  achieve best average scores. Specifically, PSGN \cite{Fan2017APS} and 3DAttriFlow \cite{Wen20223DSR}, as point-based methods, are most relevant to DiffPoint-S. These two methods follow traditional encoder-decoder pipelines to reconstruct point clouds using convolutional networks and graph neural networks respectively. In contrast, DiffPoint-S based on ViT can leverage features from input images in the same latent space, potentially improving the final reconstruction results. Consequently, DiffPoint-S is capable of predicting fine details of 3D shapes and achieving superior performance compared to its counterparts.
\par

\noindent \textbf{Qualitative comparison.}
Figure \ref{fig:single_vis} presents the qualitative results of our model and the baselines. All compared results are visualized using their publicly available implementations. As shown in Figure \ref{fig:single_vis}, most methods successfully capture the 3D geometry of the single input image. However, 3D-R2N2 produces a 3D voxel but lacks thin structures and surface details. OccNet and 3DAttriFlow can generate 3D shapes with more details and fine shapes, but they also fail to accurately reconstruct corresponding shapes. In comparison, our model excels at generating shapes with precise global features and local details, outperforming other methods. Notably, our model achieves finer detail reconstruction for jet airplanes (1st row in Figure \ref{fig:single_vis}), while the best baseline model (3DAttriFlow) struggles to recover accurate structures.

\par

\subsection{Multi View Reconstruction}

\noindent \textbf{Quantitative comparison.}
We set the number of input images to 5 for all methods compared in the multi-view reconstruction task. We implement all baselines using their public code. For voxel-based methods, we convert voxels to meshes and sample a fixed number of points from the mesh. For Pixel2Mesh++, we directly calculate CD and F-score from mesh data because its results cannot be sampled into points. As shown in Table \ref{tab:exp_multi}, our DiffPoint-M achieves state-of-the-art performance among multi-view reconstruction baselines, demonstrating its superior performance. Specifically, the average CD score is 2.10, which is much lower than tha t of 3DAttriFlow-M (2.58), and it is also the same for F-score. This result shows that DiffPoint-M can reconstruct results closer to ground truth. In terms of point-based methods, both 3DAttiFlow-M and DiffPoint-M achieve better CD scores in most categories compared to single-view reconstruction results. Additionally, due to discrepancies between generated results and publicly available ground truth data from OccNet, there are differences between single-view and multi-view CD and F-score results for 3DR2N2 and Pix2Mesh.

\par

\begin{figure*}[!htp]
\centering
\includegraphics[width=0.75\linewidth]{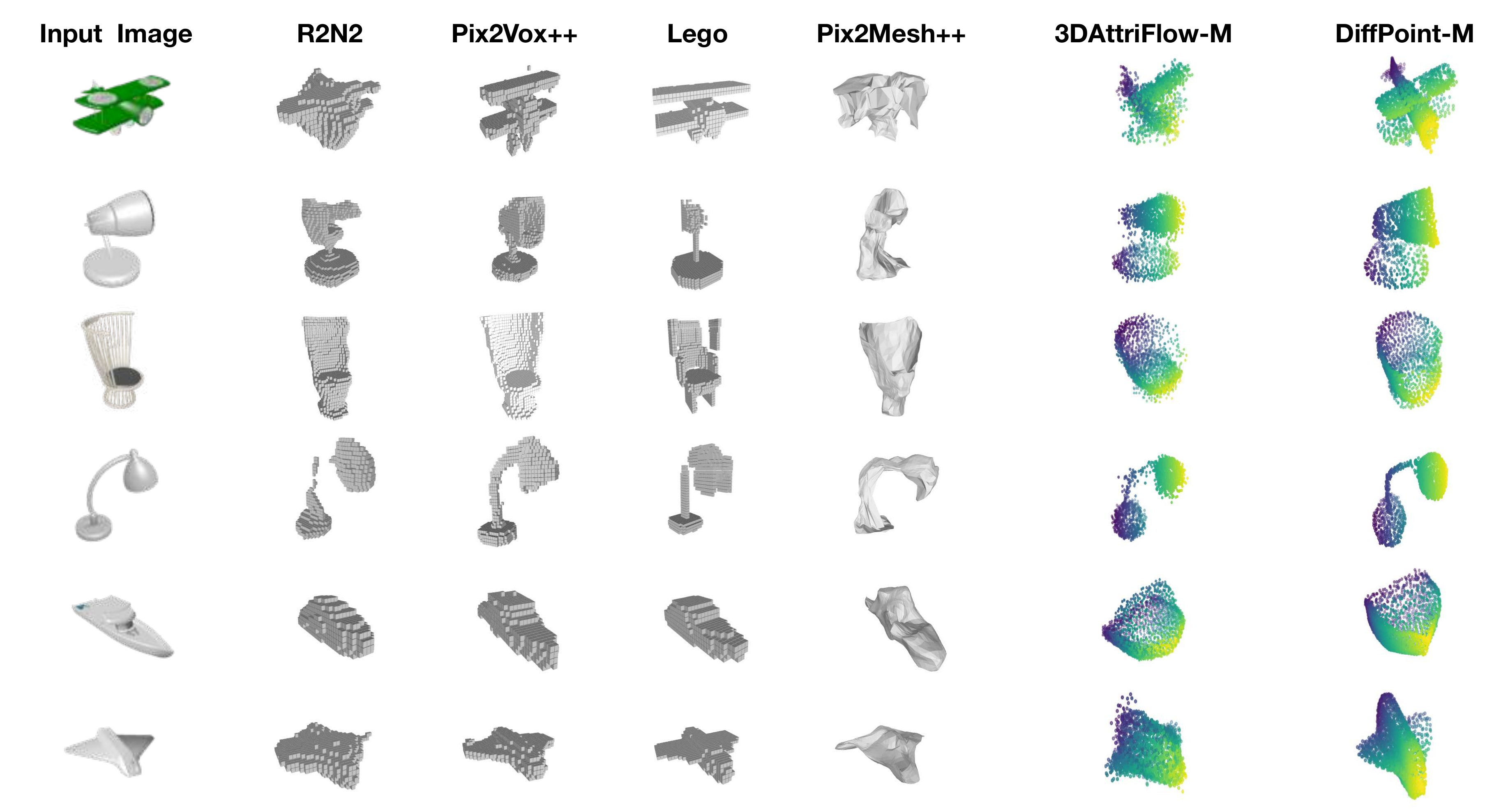}
\caption{\textbf{Multi View 3D Reconstruction.} Multi-view object reconstruction using 5 input views (only 1 is displayed). The first column shows the input image, while the remaining columns display the results of our method compared to different baselines. }
\label{fig:multi_vis}
\end{figure*}

\noindent \textbf{Qualitative comparison.}
The qualitative results of multi-view reconstruction are displayed in Figure \ref{fig:multi_vis}. Similar to the single-view results, most of the baselines can capture semantic information from the input images and reconstruct the corresponding shape. However, DiffPoint-M demonstrates superior performance in generating object details. For instance, DiffPoint-M is able to reconstruct a more accurate top shape for the lamp example, as shown in the second row of Figure \ref{fig:multi_vis}..\par
\begin{figure*}[!htp]
\centering
\includegraphics[width=0.7\linewidth]{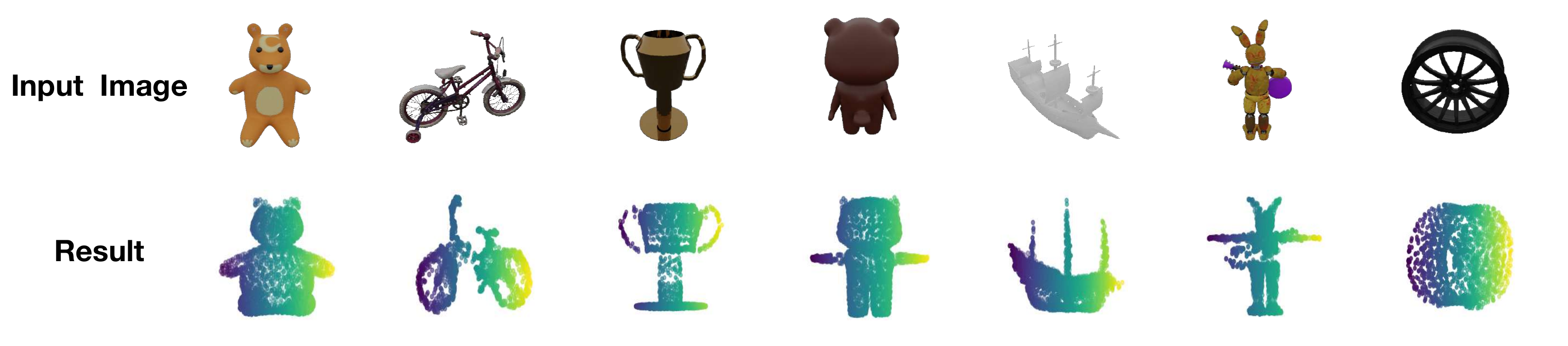}
\caption{\textbf{Multi View Reconstruction Results On OBJAVERSE-LVIS.} Models are trained using 5 input views. The first column displays one of the input images, while the second column showcases the results for our DiffPoint.}
\label{fig:lvis_vis}
\end{figure*}

Figure \ref{fig:lvis_vis} presents various reconstruction results on the OBJAVERSE-LVIS dataset. Note that we utilize all available data for training in order to showcase the effectiveness of our method in modeling complex structures. The input images used for this demonstration are selected from the training set. By observing Figure \ref{fig:lvis_vis}, it becomes evident that DiffPoint-M can be expanded to handle even more intricate structural data.

\par

\subsection{Comparison with Point Based Diffusion Model}
In this section, we compare our proposed DiffPoint model based on ViT with other point-based diffusion models in the task of single-view reconstruction: DPM \cite{Luo2021DiffusionPM} and LION \cite{Zeng2022LIONLP}. For DPM, which uses an MLP backbone, we modify its public code for single-view reconstruction by replacing their point encoder with an image encoder. We refer to this modified version as DPM-S. For LION, based on PVCNN, we follow its public implementation for single-view reconstruction and call it LION-S. In our experiment, we find that DPM-S cannot effectively model all categories of ShapeNet data. Therefore, DPM-S is trained separately for each category. It should be noted that LION-S is trained on all categories. The average CD score of 13 categories on ShapeNet is shown in table \ref{tab:exp_compare_with_point_diff}. From the results, we can conclude that DiffPoint-S performs better than DPM-S using MLP. Since LION-S focuses on reconstructing reasonable results from input images as mentioned in LION \cite{Zeng2022LIONLP}, its reconstruction results differ from the ground truth and therefore have a poor CD score. Moreover, as a latent diffusion model with 8192 dimensions of latent vectors, LION consists of two stages of training which require significant computational resources.

\begin{table}[htbp!]
\centering
\caption{Point based diffusion model comparison for single-view reconstruction on ShapeNet.}
\label{tab:exp_compare_with_point_diff}
\setlength{\tabcolsep}{15mm}{
\begin{tabular}{c| c} 
\toprule
    {Methods} &{CD$\downarrow$} \\
\midrule
      DPM-S &3.20 \\
      LION-S &8.96 \\
      \textbf{DiffPoint-S} &\textbf{2.19}\\
\bottomrule
\end{tabular}}
\end{table}

\subsection{Ablation Studies}
In this section, we validate how the key components of our model affect its performance on the multi-view 3D reconstruction task. Specifically, we analyze the effects of the multi-feature aggregation module proposed in this paper and the position embedding. All experimental setups are consistent with DiifPoint-M, except for the ablation part.

\begin{table}
    \caption{Ablation studies on ShapeNet for multi-view reconstruction task.}
    \label{tab:ablation}
    \centering
    \begin{subtable}[h]{0.495\linewidth}
        \centering
        \caption{Multi feature aggregation module}
        \label{tab:ablation_mfa}
        \begin{tabular}{cc}
            \toprule
            Module & CD$\downarrow$  \\
            \midrule
            No MFA & 2.25 \\
            With MFA & \textbf{2.10} \\
            \bottomrule
        \end{tabular}
    \end{subtable}
    \begin{subtable}[h]{0.495\linewidth}
        \centering
        \caption{Position embedding}
        \label{tab:ablation_pe}
        \begin{tabular}{cc}
            \toprule
            Module & CD$\downarrow$  \\
            \midrule
            No PE &  2.14\\
            With PE &  \textbf{2.10}\\
            \bottomrule
        \end{tabular}
    \end{subtable}
\end{table}

\noindent \textbf{Effect of Multi Feature Aggregation Module.}
To quantitatively evaluate the multi-feature aggregation module, we replace it with simple average fusion. Specifically, we refer to the model with the multi-feature aggregation module as "With MFA," otherwise known as "No MFA." Table \ref{tab:ablation_mfa} displays the average CD score for all categories on ShapeNet. The results indicate that the multi-feature aggregation model enhances the performance of our DiffPoint.

\noindent \textbf{Effect of Position Embedding.}
As 3D reconstruction task is a kind of low-level task, position embedding may play an important role in reconstruction process.  In our default setting for DiffPoint-M, we utilize a 1-dimensional learnable position embedding proposed in ViT. However, to assess the influence of position embedding, we have conducted experiments without it, referred to as "No PE". The results presented in Table \ref{tab:ablation_pe} indicate that DiffPoint-M shows minimal sensitivity towards position embedding with only a slight decrease in performance.

%% file: sec/5_conclusion.tex
\section{Conclusion}
\label{sec:conc}
In this paper, we first present DiffPoint, a novel framework of diffusion model based on ViT for point cloud reconstruction. Our DiffPoint is neat and powerful, treating all inputs including the time, image embeddings and noisy point patches as tokens. Compared with the previous methods, which cannot fully explore the visual information from images, DiffPoint takes the advantage of ViT architecture to enhance the feature representation and reduce the disparities of images and point clouds. The effectiveness of unified DiffPoint is verified on single-view and multi-view 3D reconstruction tasks. Specifically, DiffPoint outperforms other methods in terms of CD and F-score metrics for both single-view and multi-view reconstruction.

%% file: sec/X_suppl.tex
\clearpage
\setcounter{page}{1}
\setcounter{section}{0}
\renewcommand\thesection{\Alph{section}}
\maketitlesupplementary
\large

\section{Experimental Setup}
\label{sec:sup_ex_set}
\raggedright
We list the experimental setup for DiffPoint-S and DiffPoint-M presented in the main paper in Table \ref{tab:exp_setup}. In our experiments, we keep the number of groups and group size same for our three main experiments. They mainly differ from the number of blocks (depth) and the dimension of token embedding. For sampling steps, we find that 200 or 1000 steps both works well for all experiments. For fast sampling, it is better to choose 200 steps. 

\begin{table*}[htbp!]
\centering
\caption{\centering Three experimental setup for DiffPoint in the main paper. "Per category" means the model is trained on each category separately. "All categories" means the model is trained on all of categories in the dataset.}
\label{tab:exp_setup}
\scalebox{0.8}{
\begin{tabular}{c| c c c} 
\toprule[2pt]
    {Methods} &DiffPoint-S &DiffPoint-M &DiffPoint-M \\
\midrule[2pt]
\midrule[2pt]
      \#Training Params &33.06MB  &65.23MB &65.23MB\\
      Dataset &ShapeNet &ShapeNet &OBJAVERSE\\
      Training devices &1 V100 &1 V100 &1 V100\\
      Training strategy &Per category &All categories &All categories\\
      Bachsize &128 &128 &128\\
      Optimizer &AdamW  &AdamW &AdamW\\
      Learning rate &2e-4  &2e-4 &2e-4\\
      Weight decay &0.03  &0.05  &0.03 \\
      Sampling steps &200  &1000 &1000\\
      $\beta_1$ &1e-4  &1e-4  &1e-4 \\
      $\beta_T$ &0.05  &0.02 &0.02 \\
      number of input images &1  &5 &5 \\
      embedding dim &384  &512 &512 \\
      number of heads &16  &16 &16\\
      depth &16  &18 &18 \\
      number of groups &64  &64 &64 \\
      group size &32  &32 &32 \\
      drop path rate &0.1 &0.1 &0.1\\
\bottomrule[2pt]
\end{tabular}}
\end{table*}

\section{Additional Samples}
\begin{figure*}[!htbp]
\centering
\includegraphics[width=0.7\linewidth]{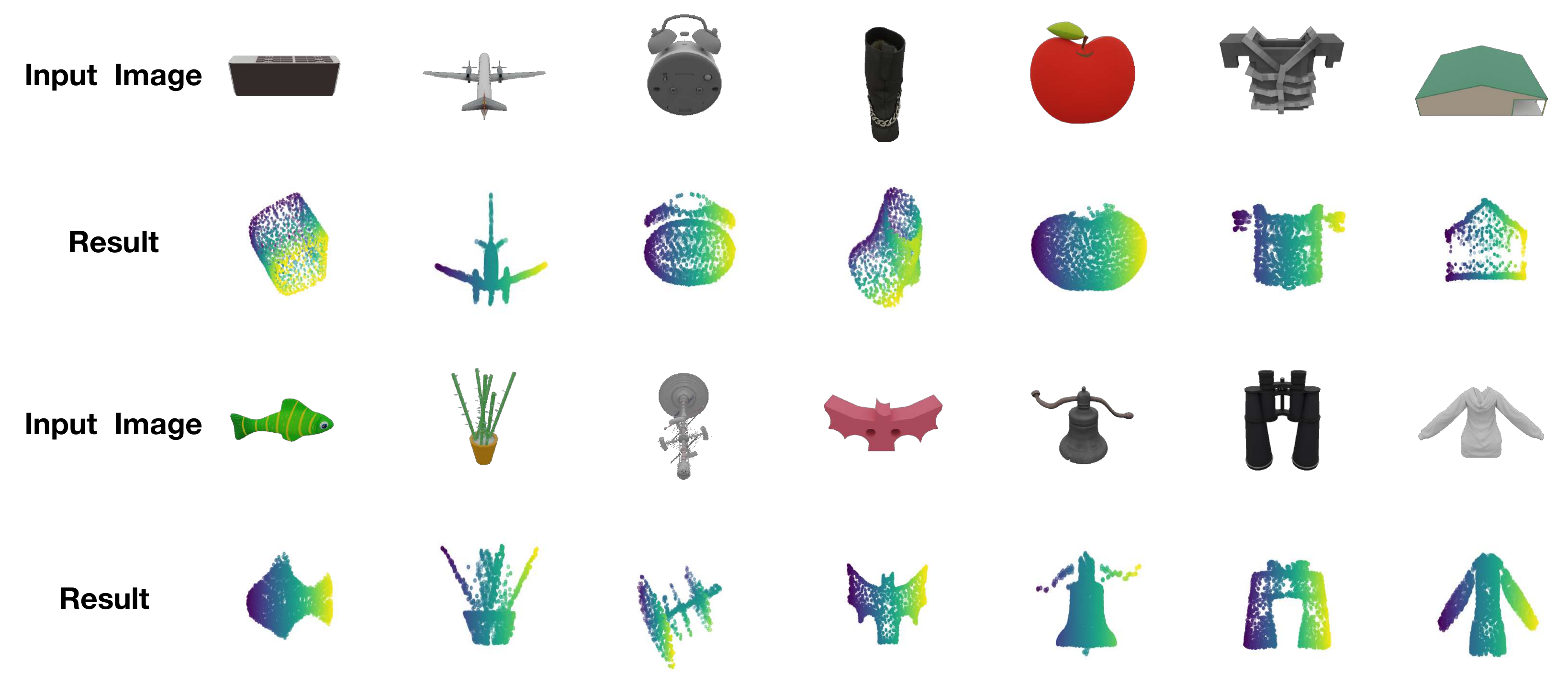}
\caption{Generated samples on OBJAVERSE-LVIS dataset.}
\label{fig:lvis_supp}
\end{figure*}

%% file: main.bbl
\begin{thebibliography}{65}
\providecommand{\natexlab}[1]{#1}
\providecommand{\url}[1]{\texttt{#1}}
\expandafter\ifx\csname urlstyle\endcsname\relax
  \providecommand{\doi}[1]{doi: #1}\else
  \providecommand{\doi}{doi: \begingroup \urlstyle{rm}\Url}\fi

\bibitem[Bao et~al.(2022)Bao, Nie, Xue, Cao, Li, Su, and Zhu]{Bao2022AllAW}
Fan Bao, Shen Nie, Kaiwen Xue, Yue Cao, Chongxuan Li, Hang Su, and Jun Zhu.
\newblock All are worth words: A vit backbone for diffusion models.
\newblock \emph{2023 IEEE/CVF Conference on Computer Vision and Pattern Recognition (CVPR)}, pages 22669--22679, 2022.

\bibitem[Barron et~al.(2021)Barron, Mildenhall, Tancik, Hedman, Martin-Brualla, and Srinivasan]{Barron2021MipNeRFAM}
Jonathan~T. Barron, Ben Mildenhall, Matthew Tancik, Peter Hedman, Ricardo Martin-Brualla, and Pratul~P. Srinivasan.
\newblock Mip-nerf: A multiscale representation for anti-aliasing neural radiance fields.
\newblock \emph{2021 IEEE/CVF International Conference on Computer Vision (ICCV)}, pages 5835--5844, 2021.

\bibitem[Bian et~al.(2022)Bian, Wang, Li, Bian, and Prisacariu]{Bian2022NoPeNeRFON}
Wenjing Bian, Zirui Wang, Kejie Li, Jiawang Bian, and Victor~Adrian Prisacariu.
\newblock Nope-nerf: Optimising neural radiance field with no pose prior.
\newblock \emph{2023 IEEE/CVF Conference on Computer Vision and Pattern Recognition (CVPR)}, pages 4160--4169, 2022.

\bibitem[Brown et~al.(2020)Brown, Mann, Ryder, Subbiah, Kaplan, Dhariwal, Neelakantan, Shyam, Sastry, Askell, Agarwal, Herbert-Voss, Krueger, Henighan, Child, Ramesh, Ziegler, Wu, Winter, Hesse, Chen, Sigler, Litwin, Gray, Chess, Clark, Berner, McCandlish, Radford, Sutskever, and Amodei]{Brown2020LanguageMA}
Tom~B. Brown, Benjamin Mann, Nick Ryder, Melanie Subbiah, Jared Kaplan, Prafulla Dhariwal, Arvind Neelakantan, Pranav Shyam, Girish Sastry, Amanda Askell, Sandhini Agarwal, Ariel Herbert-Voss, Gretchen Krueger, T.~J. Henighan, Rewon Child, Aditya Ramesh, Daniel~M. Ziegler, Jeff Wu, Clemens Winter, Christopher Hesse, Mark Chen, Eric Sigler, Mateusz Litwin, Scott Gray, Benjamin Chess, Jack Clark, Christopher Berner, Sam McCandlish, Alec Radford, Ilya Sutskever, and Dario Amodei.
\newblock Language models are few-shot learners.
\newblock \emph{ArXiv}, abs/2005.14165, 2020.

\bibitem[Chang et~al.(2015)Chang, Funkhouser, Guibas, Hanrahan, Huang, Li, Savarese, Savva, Song, Su, Xiao, Yi, and Yu]{Chang2015ShapeNetAI}
Angel~X. Chang, Thomas~A. Funkhouser, Leonidas~J. Guibas, Pat Hanrahan, Qixing Huang, Zimo Li, Silvio Savarese, Manolis Savva, Shuran Song, Hao Su, Jianxiong Xiao, L. Yi, and Fisher Yu.
\newblock Shapenet: An information-rich 3d model repository.
\newblock \emph{ArXiv}, abs/1512.03012, 2015.

\bibitem[Chang et~al.(2023)Chang, Zhang, Barber, Maschinot, Lezama, Jiang, Yang, Murphy, Freeman, Rubinstein, Li, and Krishnan]{Chang2023MuseTG}
Huiwen Chang, Han Zhang, Jarred Barber, AJ Maschinot, Jos{\'e} Lezama, Lu Jiang, Ming Yang, Kevin~P. Murphy, William~T. Freeman, Michael Rubinstein, Yuanzhen Li, and Dilip Krishnan.
\newblock Muse: Text-to-image generation via masked generative transformers.
\newblock \emph{ArXiv}, abs/2301.00704, 2023.

\bibitem[Choy et~al.(2016)Choy, Xu, Gwak, Chen, and Savarese]{Choy20163DR2N2AU}
Christopher~Bongsoo Choy, Danfei Xu, JunYoung Gwak, Kevin Chen, and Silvio Savarese.
\newblock 3d-r2n2: A unified approach for single and multi-view 3d object reconstruction.
\newblock In \emph{ECCV}, 2016.

\bibitem[Deitke et~al.(2022)Deitke, Schwenk, Salvador, Weihs, Michel, VanderBilt, Schmidt, Ehsani, Kembhavi, and Farhadi]{Deitke2022ObjaverseAU}
Matt Deitke, Dustin Schwenk, Jordi Salvador, Luca Weihs, Oscar Michel, Eli VanderBilt, Ludwig Schmidt, Kiana Ehsani, Aniruddha Kembhavi, and Ali Farhadi.
\newblock Objaverse: A universe of annotated 3d objects.
\newblock \emph{ArXiv}, abs/2212.08051, 2022.

\bibitem[Devlin et~al.(2019)Devlin, Chang, Lee, and Toutanova]{Devlin2019BERTPO}
Jacob Devlin, Ming-Wei Chang, Kenton Lee, and Kristina Toutanova.
\newblock Bert: Pre-training of deep bidirectional transformers for language understanding.
\newblock In \emph{North American Chapter of the Association for Computational Linguistics}, 2019.

\bibitem[Dhariwal and Nichol(2021)]{Dhariwal2021DiffusionMB}
Prafulla Dhariwal and Alex Nichol.
\newblock Diffusion models beat gans on image synthesis.
\newblock \emph{NeurIPS}, 2021.

\bibitem[Dosovitskiy et~al.(2020)Dosovitskiy, Beyer, Kolesnikov, Weissenborn, Zhai, Unterthiner, Dehghani, Minderer, Heigold, Gelly, Uszkoreit, and Houlsby]{Dosovitskiy2020AnII}
Alexey Dosovitskiy, Lucas Beyer, Alexander Kolesnikov, Dirk Weissenborn, Xiaohua Zhai, Thomas Unterthiner, Mostafa Dehghani, Matthias Minderer, Georg Heigold, Sylvain Gelly, Jakob Uszkoreit, and Neil Houlsby.
\newblock An image is worth 16x16 words: Transformers for image recognition at scale.
\newblock \emph{ArXiv}, abs/2010.11929, 2020.

\bibitem[Engel et~al.(2020)Engel, Belagiannis, and Dietmayer]{Engel2020PointT}
Nico Engel, Vasileios Belagiannis, and Klaus C.~J. Dietmayer.
\newblock Point transformer.
\newblock \emph{IEEE Access}, 9:\penalty0 134826--134840, 2020.

\bibitem[Fan et~al.(2017)Fan, Su, and Guibas]{Fan2017APS}
Haoqiang Fan, Hao Su, and Leonidas~J. Guibas.
\newblock A point set generation network for 3d object reconstruction from a single image.
\newblock \emph{2017 IEEE Conference on Computer Vision and Pattern Recognition (CVPR)}, pages 2463--2471, 2017.

\bibitem[Groueix et~al.(2018)Groueix, Fisher, Kim, Russell, and Aubry]{Groueix2018AtlasNetAP}
Thibault Groueix, Matthew Fisher, Vladimir~G. Kim, Bryan~C. Russell, and Mathieu Aubry.
\newblock Atlasnet: A papier-m\^ach\'e approach to learning 3d surface generation.
\newblock \emph{CVPR}, 2018.

\bibitem[He et~al.(2021)He, Chen, Xie, Li, Doll'ar, and Girshick]{He2021MaskedAA}
Kaiming He, Xinlei Chen, Saining Xie, Yanghao Li, Piotr Doll'ar, and Ross~B. Girshick.
\newblock Masked autoencoders are scalable vision learners.
\newblock \emph{2022 IEEE/CVF Conference on Computer Vision and Pattern Recognition (CVPR)}, pages 15979--15988, 2021.

\bibitem[Ho et~al.(2020)Ho, Jain, and Abbeel]{Ho2020DenoisingDP}
Jonathan Ho, Ajay Jain, and P. Abbeel.
\newblock Denoising diffusion probabilistic models.
\newblock \emph{NeurIPS}, 2020.

\bibitem[Huang et~al.(2022)Huang, Shi, Li, Li, Huang, and Li]{Huang2022MultimodalSF}
Keli Huang, Botian Shi, Xiang Li, Xin Li, Siyuan Huang, and Yikang Li.
\newblock Multi-modal sensor fusion for auto driving perception: A survey.
\newblock \emph{ArXiv}, abs/2202.02703, 2022.

\bibitem[Kim et~al.(2022)Kim, Kim, and Yoon]{Kim2022GuidedTTSAD}
Heeseung Kim, Sungwon Kim, and Sungroh Yoon.
\newblock Guided-tts: A diffusion model for text-to-speech via classifier guidance.
\newblock In \emph{ICML}, 2022.

\bibitem[Kong et~al.(2021)Kong, Ping, Huang, Zhao, and Catanzaro]{Kong2021DiffWaveAV}
Zhifeng Kong, Wei Ping, Jiaji Huang, Kexin Zhao, and Bryan Catanzaro.
\newblock Diffwave: A versatile diffusion model for audio synthesis.
\newblock \emph{ICLR}, 2021.

\bibitem[Kumar et~al.(2022)Kumar, Eising, Witt, and Yogamani]{Kumar2022SurroundviewFC}
Varun~Ravi Kumar, Ciar{\'a}n Eising, Christian Witt, and Senthil~Kumar Yogamani.
\newblock Surround-view fisheye camera perception for automated driving: Overview, survey and challenges.
\newblock \emph{ArXiv}, abs/2205.13281, 2022.

\bibitem[Liu et~al.(2022)Liu, Cai, and Lee]{Liu2022MaskedDF}
Haotian Liu, Mu Cai, and Yong~Jae Lee.
\newblock Masked discrimination for self-supervised learning on point clouds.
\newblock In \emph{European Conference on Computer Vision}, 2022.

\bibitem[Liu et~al.(2020)Liu, Gu, Lin, Chua, and Theobalt]{Liu2020NeuralSV}
Lingjie Liu, Jiatao Gu, Kyaw~Zaw Lin, Tat-Seng Chua, and Christian Theobalt.
\newblock Neural sparse voxel fields.
\newblock \emph{ArXiv}, abs/2007.11571, 2020.

\bibitem[Liu et~al.(2021)Liu, Lin, Cao, Hu, Wei, Zhang, Lin, and Guo]{Liu2021SwinTH}
Ze Liu, Yutong Lin, Yue Cao, Han Hu, Yixuan Wei, Zheng Zhang, Stephen Lin, and Baining Guo.
\newblock Swin transformer: Hierarchical vision transformer using shifted windows.
\newblock \emph{2021 IEEE/CVF International Conference on Computer Vision (ICCV)}, pages 9992--10002, 2021.

\bibitem[Loshchilov and Hutter(2017)]{Loshchilov2017DecoupledWD}
Ilya Loshchilov and Frank Hutter.
\newblock Decoupled weight decay regularization.
\newblock In \emph{International Conference on Learning Representations}, 2017.

\bibitem[Lunghi et~al.(2017)Lunghi, Prades, Castro, Ferre, and Masi]{Lunghi2017AnRB}
Giacomo Lunghi, Ra{\'u}l~Mar{\'i}n Prades, Mario~Di Castro, Manuel Ferre, and Alessandro Masi.
\newblock An rgb-d based augmented reality 3d reconstruction system for robotic environmental inspection of radioactive areas.
\newblock In \emph{ICINCO}, 2017.

\bibitem[Luo and Hu(2021)]{Luo2021DiffusionPM}
Shitong Luo and Wei Hu.
\newblock Diffusion probabilistic models for 3d point cloud generation.
\newblock \emph{2021 IEEE/CVF Conference on Computer Vision and Pattern Recognition (CVPR)}, pages 2836--2844, 2021.

\bibitem[Maboudi et~al.(2022)Maboudi, Homaei, Song, Malihi, Saadatseresht, and Gerke]{Maboudi2022ARO}
Mehdi Maboudi, Mohammad~Hossein Homaei, Soohwan Song, Shirin Malihi, Mohammad Saadatseresht, and Markus Gerke.
\newblock A review on viewpoints and path-planning for uav-based 3d reconstruction.
\newblock \emph{ArXiv}, abs/2205.03716, 2022.

\bibitem[Mescheder et~al.(2019)Mescheder, Oechsle, Niemeyer, Nowozin, and Geiger]{Mescheder2019OccupancyNL}
Lars~M. Mescheder, Michael Oechsle, Michael Niemeyer, Sebastian Nowozin, and Andreas Geiger.
\newblock Occupancy networks: Learning 3d reconstruction in function space.
\newblock \emph{2019 IEEE/CVF Conference on Computer Vision and Pattern Recognition (CVPR)}, pages 4455--4465, 2019.

\bibitem[Mildenhall et~al.(2020)Mildenhall, Srinivasan, Tancik, Barron, Ramamoorthi, and Ng]{Mildenhall2020NeRF}
Ben Mildenhall, Pratul~P. Srinivasan, Matthew Tancik, Jonathan~T. Barron, Ravi Ramamoorthi, and Ren Ng.
\newblock Nerf.
\newblock \emph{Communications of the ACM}, 65:\penalty0 99 -- 106, 2020.

\bibitem[Mo et~al.(2023)Mo, Xie, Chu, Yao, Hong, Nie{\ss}ner, and Li]{Mo2023DiT3DEP}
Shentong Mo, Enze Xie, Ruihang Chu, Lewei Yao, Lanqing Hong, Matthias Nie{\ss}ner, and Zhenguo Li.
\newblock Dit-3d: Exploring plain diffusion transformers for 3d shape generation.
\newblock \emph{ArXiv}, abs/2307.01831, 2023.

\bibitem[Muller et~al.(2022)Muller, Siddiqui, Porzi, Bul{\`o}, Kontschieder, and Nie{\ss}ner]{Muller2022DiffRFR3}
Norman Muller, Yawar Siddiqui, Lorenzo Porzi, Samuel~Rota Bul{\`o}, Peter Kontschieder, and Matthias Nie{\ss}ner.
\newblock Diffrf: Rendering-guided 3d radiance field diffusion.
\newblock \emph{2023 IEEE/CVF Conference on Computer Vision and Pattern Recognition (CVPR)}, pages 4328--4338, 2022.

\bibitem[Nichol et~al.(2022)Nichol, Dhariwal, Ramesh, Shyam, Mishkin, McGrew, Sutskever, and Chen]{Nichol2022GLIDETP}
Alex Nichol, Prafulla Dhariwal, Aditya Ramesh, Pranav Shyam, Pamela Mishkin, Bob McGrew, Ilya Sutskever, and Mark Chen.
\newblock Glide: Towards photorealistic image generation and editing with text-guided diffusion models.
\newblock In \emph{ICML}, 2022.

\bibitem[Pan et~al.(2019)Pan, Han, Chen, Tang, and Jia]{Pan2019DeepMR}
Junyi Pan, Xiaoguang Han, Weikai Chen, Jiapeng Tang, and Kui Jia.
\newblock Deep mesh reconstruction from single rgb images via topology modification networks.
\newblock \emph{2019 IEEE/CVF International Conference on Computer Vision (ICCV)}, pages 9963--9972, 2019.

\bibitem[Pang et~al.(2022)Pang, Wang, Tay, Liu, Tian, and Yuan]{Pang2022MaskedAF}
Yatian Pang, Wenxiao Wang, Francis E.~H. Tay, W. Liu, Yonghong Tian, and Liuliang Yuan.
\newblock Masked autoencoders for point cloud self-supervised learning.
\newblock In \emph{European Conference on Computer Vision}, 2022.

\bibitem[Paszke et~al.(2017)Paszke, Gross, Chintala, Chanan, Yang, DeVito, Lin, Desmaison, Antiga, and Lerer]{Paszke2017AutomaticDI}
Adam Paszke, Sam Gross, Soumith Chintala, Gregory Chanan, Edward Yang, Zach DeVito, Zeming Lin, Alban Desmaison, Luca Antiga, and Adam Lerer.
\newblock Automatic differentiation in pytorch.
\newblock 2017.

\bibitem[Popov et~al.(2021)Popov, Vovk, Gogoryan, Sadekova, and Kudinov]{Popov2021GradTTSAD}
Vadim Popov, Ivan Vovk, Vladimir Gogoryan, Tasnima Sadekova, and Mikhail Kudinov.
\newblock Grad-tts: A diffusion probabilistic model for text-to-speech.
\newblock \emph{ICML}, 2021.

\bibitem[Preechakul et~al.(2021)Preechakul, Chatthee, Wizadwongsa, and Suwajanakorn]{Preechakul2021DiffusionAT}
Konpat Preechakul, Nattanat Chatthee, Suttisak Wizadwongsa, and Supasorn Suwajanakorn.
\newblock Diffusion autoencoders: Toward a meaningful and decodable representation.
\newblock \emph{2022 IEEE/CVF Conference on Computer Vision and Pattern Recognition (CVPR)}, pages 10609--10619, 2021.

\bibitem[Qi et~al.(2016)Qi, Su, Mo, and Guibas]{Qi2016PointNetDL}
C. Qi, Hao Su, Kaichun Mo, and Leonidas~J. Guibas.
\newblock Pointnet: Deep learning on point sets for 3d classification and segmentation.
\newblock \emph{2017 IEEE Conference on Computer Vision and Pattern Recognition (CVPR)}, pages 77--85, 2016.

\bibitem[Qi et~al.(2017)Qi, Yi, Su, and Guibas]{Qi2017PointNetDH}
C. Qi, L. Yi, Hao Su, and Leonidas~J. Guibas.
\newblock Pointnet++: Deep hierarchical feature learning on point sets in a metric space.
\newblock In \emph{Neural Information Processing Systems}, 2017.

\bibitem[Radford and Narasimhan(2018)]{Radford2018ImprovingLU}
Alec Radford and Karthik Narasimhan.
\newblock Improving language understanding by generative pre-training.
\newblock 2018.

\bibitem[Radford et~al.(2021)Radford, Kim, Hallacy, Ramesh, Goh, Agarwal, Sastry, Askell, Mishkin, Clark, Krueger, and Sutskever]{Radford2021LearningTV}
Alec Radford, Jong~Wook Kim, Chris Hallacy, Aditya Ramesh, Gabriel Goh, Sandhini Agarwal, Girish Sastry, Amanda Askell, Pamela Mishkin, Jack Clark, Gretchen Krueger, and Ilya Sutskever.
\newblock Learning transferable visual models from natural language supervision.
\newblock In \emph{International Conference on Machine Learning}, 2021.

\bibitem[Raffel et~al.(2019)Raffel, Shazeer, Roberts, Lee, Narang, Matena, Zhou, Li, and Liu]{Raffel2019ExploringTL}
Colin Raffel, Noam~M. Shazeer, Adam Roberts, Katherine Lee, Sharan Narang, Michael Matena, Yanqi Zhou, Wei Li, and Peter~J. Liu.
\newblock Exploring the limits of transfer learning with a unified text-to-text transformer.
\newblock \emph{J. Mach. Learn. Res.}, 21:\penalty0 140:1--140:67, 2019.

\bibitem[Ramesh et~al.(2022)Ramesh, Dhariwal, Nichol, Chu, and Chen]{Ramesh2022HierarchicalTI}
Aditya Ramesh, Prafulla Dhariwal, Alex Nichol, Casey Chu, and Mark Chen.
\newblock Hierarchical text-conditional image generation with clip latents.
\newblock \emph{ArXiv}, abs/2204.06125, 2022.

\bibitem[Rombach et~al.(2021)Rombach, Blattmann, Lorenz, Esser, and Ommer]{Rombach2021HighResolutionIS}
Robin Rombach, A. Blattmann, Dominik Lorenz, Patrick Esser, and Bj{\"o}rn Ommer.
\newblock High-resolution image synthesis with latent diffusion models.
\newblock \emph{2022 IEEE/CVF Conference on Computer Vision and Pattern Recognition (CVPR)}, pages 10674--10685, 2021.

\bibitem[Sohl-Dickstein et~al.(2015)Sohl-Dickstein, Weiss, Maheswaranathan, and Ganguli]{SohlDickstein2015DeepUL}
Jascha~Narain Sohl-Dickstein, Eric~A. Weiss, Niru Maheswaranathan, and Surya Ganguli.
\newblock Deep unsupervised learning using nonequilibrium thermodynamics.
\newblock \emph{ICML}, 2015.

\bibitem[Song et~al.(2020)Song, Sohl-Dickstein, Kingma, Kumar, Ermon, and Poole]{Song2020ScoreBasedGM}
Yang Song, Jascha~Narain Sohl-Dickstein, Diederik~P. Kingma, Abhishek Kumar, Stefano Ermon, and Ben Poole.
\newblock Score-based generative modeling through stochastic differential equations.
\newblock \emph{ArXiv}, abs/2011.13456, 2020.

\bibitem[Tashiro et~al.(2021)Tashiro, Song, Song, and Ermon]{Tashiro2021CSDICS}
Yusuke Tashiro, Jiaming Song, Yang Song, and Stefano Ermon.
\newblock Csdi: Conditional score-based diffusion models for probabilistic time series imputation.
\newblock In \emph{NeurIPS}, 2021.

\bibitem[Tatarchenko et~al.(2019)Tatarchenko, Richter, Ranftl, Li, Koltun, and Brox]{Tatarchenko2019WhatDS}
Maxim Tatarchenko, Stephan~R. Richter, Ren{\'e} Ranftl, Zhuwen Li, Vladlen Koltun, and Thomas Brox.
\newblock What do single-view 3d reconstruction networks learn?
\newblock \emph{2019 IEEE/CVF Conference on Computer Vision and Pattern Recognition (CVPR)}, pages 3400--3409, 2019.

\bibitem[Vaswani et~al.(2017)Vaswani, Shazeer, Parmar, Uszkoreit, Jones, Gomez, Kaiser, and Polosukhin]{Vaswani2017AttentionIA}
Ashish Vaswani, Noam~M. Shazeer, Niki Parmar, Jakob Uszkoreit, Llion Jones, Aidan~N. Gomez, Lukasz Kaiser, and Illia Polosukhin.
\newblock Attention is all you need.
\newblock In \emph{NIPS}, 2017.

\bibitem[Wang et~al.(2018{\natexlab{a}})Wang, Zhang, Li, Fu, Liu, and Jiang]{Wang2018Pixel2MeshG3}
Nanyang Wang, Yinda Zhang, Zhuwen Li, Yanwei Fu, W. Liu, and Yu-Gang Jiang.
\newblock Pixel2mesh: Generating 3d mesh models from single rgb images.
\newblock In \emph{ECCV}, 2018{\natexlab{a}}.

\bibitem[Wang et~al.(2018{\natexlab{b}})Wang, Sun, Liu, and Tong]{Wang2018AdaptiveO}
Peng-Shuai Wang, Chun-Yu Sun, Yang Liu, and Xin Tong.
\newblock Adaptive o-cnn.
\newblock \emph{ACM Transactions on Graphics (TOG)}, 37:\penalty0 1 -- 11, 2018{\natexlab{b}}.

\bibitem[Wen et~al.(2019)Wen, Zhang, Li, and Fu]{Wen2019Pixel2MeshM3}
Chao Wen, Yinda Zhang, Zhuwen Li, and Yanwei Fu.
\newblock Pixel2mesh++: Multi-view 3d mesh generation via deformation.
\newblock \emph{2019 IEEE/CVF International Conference on Computer Vision (ICCV)}, pages 1042--1051, 2019.

\bibitem[Wen et~al.(2022{\natexlab{a}})Wen, Zhang, Cao, Li, Xue, and Fu]{Wen2022Pixel2Mesh3M}
Chao Wen, Yinda Zhang, Chenjie Cao, Zhuwen Li, X. Xue, and Yanwei Fu.
\newblock Pixel2mesh++: 3d mesh generation and refinement from multi-view images.
\newblock \emph{IEEE Transactions on Pattern Analysis and Machine Intelligence}, 45:\penalty0 2166--2180, 2022{\natexlab{a}}.

\bibitem[Wen et~al.(2022{\natexlab{b}})Wen, Zhou, Liu, Dong, and Han]{Wen20223DSR}
Xin Wen, Junsheng Zhou, Yu-Shen Liu, Zhen Dong, and Zhizhong Han.
\newblock 3d shape reconstruction from 2d images with disentangled attribute flow.
\newblock \emph{2022 IEEE/CVF Conference on Computer Vision and Pattern Recognition (CVPR)}, pages 3793--3803, 2022{\natexlab{b}}.

\bibitem[Wu et~al.(2015)Wu, Song, Khosla, Yu, Zhang, Tang, and Xiao]{Wu20153DSA}
Zhirong Wu, Shuran Song, Aditya Khosla, Fisher Yu, Linguang Zhang, Xiaoou Tang, and Jianxiong Xiao.
\newblock 3d shapenets: A deep representation for volumetric shapes.
\newblock \emph{2015 IEEE Conference on Computer Vision and Pattern Recognition (CVPR)}, pages 1912--1920, 2015.

\bibitem[Wynn and Turmukhambetov(2023)]{Wynn2023DiffusioNeRFRN}
Jamie~M. Wynn and Daniyar Turmukhambetov.
\newblock Diffusionerf: Regularizing neural radiance fields with denoising diffusion models.
\newblock \emph{2023 IEEE/CVF Conference on Computer Vision and Pattern Recognition (CVPR)}, pages 4180--4189, 2023.

\bibitem[Xie et~al.(2019)Xie, Yao, Sun, Zhou, Zhang, and Tong]{Xie2019Pix2VoxC3}
Haozhe Xie, Hongxun Yao, Xiaoshuai Sun, Shangchen Zhou, Shengping Zhang, and Xiaojun Tong.
\newblock Pix2vox: Context-aware 3d reconstruction from single and multi-view images.
\newblock \emph{2019 IEEE/CVF International Conference on Computer Vision (ICCV)}, pages 2690--2698, 2019.

\bibitem[Xie et~al.(2020)Xie, Yao, Zhang, Zhou, and Sun]{Xie2020Pix2VoxMC}
Haozhe Xie, Hongxun Yao, Shengping Zhang, Shangchen Zhou, and Wenxiu Sun.
\newblock Pix2vox++: Multi-scale context-aware 3d object reconstruction from single and multiple images.
\newblock \emph{International Journal of Computer Vision}, pages 1 -- 17, 2020.

\bibitem[Yagubbayli et~al.(2021)Yagubbayli, Tonioni, and Tombari]{Yagubbayli2021LegoFormerTF}
Farid Yagubbayli, Alessio Tonioni, and Federico Tombari.
\newblock Legoformer: Transformers for block-by-block multi-view 3d reconstruction.
\newblock \emph{CVPR}, abs/2106.12102, 2021.

\bibitem[Yang et~al.(2018)Yang, Wang, Markham, and Trigoni]{Yang2018RobustAA}
Bo Yang, Sen Wang, A. Markham, and Niki Trigoni.
\newblock Robust attentional aggregation of deep feature sets for multi-view 3d reconstruction.
\newblock \emph{International Journal of Computer Vision}, 128:\penalty0 53 -- 73, 2018.

\bibitem[Yu et~al.(2020)Yu, Ye, Tancik, and Kanazawa]{Yu2020pixelNeRFNR}
Alex Yu, Vickie Ye, Matthew Tancik, and Angjoo Kanazawa.
\newblock pixelnerf: Neural radiance fields from one or few images.
\newblock \emph{2021 IEEE/CVF Conference on Computer Vision and Pattern Recognition (CVPR)}, pages 4576--4585, 2020.

\bibitem[Yu et~al.(2021)Yu, Tang, Rao, Huang, Zhou, and Lu]{Yu2021PointBERTP3}
Xumin Yu, Lulu Tang, Yongming Rao, Tiejun Huang, Jie Zhou, and Jiwen Lu.
\newblock Point-bert: Pre-training 3d point cloud transformers with masked point modeling.
\newblock \emph{2022 IEEE/CVF Conference on Computer Vision and Pattern Recognition (CVPR)}, pages 19291--19300, 2021.

\bibitem[Yuan et~al.(2021)Yuan, Chen, Wang, Yu, Shi, Tay, Feng, and Yan]{Yuan2021TokenstoTokenVT}
Li Yuan, Yunpeng Chen, Tao Wang, Weihao Yu, Yujun Shi, Francis E.~H. Tay, Jiashi Feng, and Shuicheng Yan.
\newblock Tokens-to-token vit: Training vision transformers from scratch on imagenet.
\newblock \emph{2021 IEEE/CVF International Conference on Computer Vision (ICCV)}, pages 538--547, 2021.

\bibitem[Zeng et~al.(2022)Zeng, Vahdat, Williams, Gojcic, Litany, Fidler, and Kreis]{Zeng2022LIONLP}
Xiaohui Zeng, Arash Vahdat, Francis Williams, Zan Gojcic, Or Litany, Sanja Fidler, and Karsten Kreis.
\newblock Lion: Latent point diffusion models for 3d shape generation.
\newblock \emph{ArXiv}, abs/2210.06978, 2022.

\bibitem[Zhou et~al.(2021)Zhou, Du, and Wu]{Zhou20213DSG}
Linqi Zhou, Yilun Du, and Jiajun Wu.
\newblock 3d shape generation and completion through point-voxel diffusion.
\newblock \emph{2021 IEEE/CVF International Conference on Computer Vision (ICCV)}, pages 5806--5815, 2021.

\end{thebibliography}
